# Condition Monitoring of HV Bushings in the Presence of Missing Data Using Evolutionary Computing

**SIZWE M. DHLAMINI**[*], **FULUFHELO V. NELWAMONDO**[**], **TSHILIDZI MARWALA**[**]

[*]Eskom, Distribution Technology
Private Bag x1074, Germiston, 1400
SOUTH AFRICA
sizwe.dhlamini@eskom.co.za

[**]School of Electrical and Information Engineering
University of the Witwatersrand, Private Bag x3,
Johannesburg, 2000,
SOUTH AFRICA
t.marwala@ee.wits.ac.za, vnelwamondo@yahoo.com
http://dept.ee.wits.ac.za/~marwala/

*Abstract:* - The work proposes the application of neural networks with particle swarm optimisation (PSO) and genetic algorithms (GA) to compensate for missing data in classifying high voltage bushings. The classification is done using DGA data from 60966 bushings based on IEEEc57.104, IEC599 and IEEE production rates methods for oil impregnated paper (OIP) bushings. PSO and GA were compared in terms of accuracy and computational efficiency. Both GA and PSO simulations were able to estimate missing data values to an average 95% accuracy when only one variable was missing. However PSO rapidly deteriorated to 66% accuracy with two variables missing simultaneously, compared to 84% for GA. The data estimated using GA was found to classify the conditions of bushings than the PSO.

## 1 Introduction

This work investigates tools that compensate for sensor failure in systems that are used for condition monitoring of high voltage bushings. Sensor failure is a concern that needs to be taken into account when designing an automated system of bushing condition monitoring. If a sensor fails on an online bushing monitoring system and the system trips a transformer, the financial and legal consequences can be serious. A failed sensor may take a few hours or months to repair so one will need to answer the following questions before implementing an online diagnostics system. What happens if one or more of the sensors fail? How many sensors can fail before the online system is rendered ineffective?

Recognising the value of artificial intelligence and the risks associated with its usage, Cigre formed Study Committee 15 WG 11 in 2002 with the task of standardising and improving applications of data mining techniques within power systems [1]. McGrail, *et al.* [1] identified several areas in reliability centred maintenance that can be improved by using artificial intelligence online and offline. Artificial intelligence (AI) can add significant value in power systems operations, maintenance and control, by providing a single engine to execute the critical role of data-fusion [2]. However, it is also clear that AI systems need to be robust and reliable.

Some of the methods used previously to account for missing data include regression techniques by Jackson [3] as well as Madow, *et al.* [4] who used principal component analysis. Abdella and Marwala [5] proposed a method of accurately compensating for missing data using an autoencoder. Markey and Patel [6] chose to use zeros where there was missing data. Other work by Ghahramani [7] and Tresp [8] also used regression to address the problem of missing data. The accuracy of the solution derived by replacing missing data with averages or zeros or an iterated number depends on whether the final diagnosis decision is based on the missing variable alone, or if that decision depends collectively on all the variables. The accuracy of the approximated variable will be considered in this paper.

What is important at all stages in evaluating data is to note that data or measurements or numbers by themselves have little meaning, and historically human intervention is required before decisions can be made and executed. Artificial Intelligence gives the option to autonomously transform, correlate and interpret data, so that it becomes valuable knowledge.

## 2 Why Missing Data is a Problem

The first reason for concern when a sensor fails is that no information is available for a particular measured parameter, and that missing variable might be important. The second reason to be concerned is that the processor of the online diagnostics tool will identify an undefined value where there is missing data due to sensor failure. The third reason for concern is that data from a sensor can become corrupted due to loss of calibration or excessive noise. The problem is that analytical tools such as neural network, fuzzy set theory, principal component analysis, etc., cannot process undefined values. Currently there is no method for specifying missing data within a neural network or fuzzy or toolbox, so missing data must be approximated prior to processing. For multivariate data, five methods are available to address missing data, namely: (1) average values of previous values of that variable; (2) average values of the training set data for that variable; (3) using zero where there is missing data; (4) deleting the variables that are undefined; and (5) finding a correlation between the missing variable and the remaining variables. Using a constant, such as averages or zeros or deleting the variable completely, provides incorrect solutions, so these options cannot be applied as a generic solution to missing data problems. Iterative methods that reduce dimensionality of data can be used to look for correlation between the measured data. This work approximates data which missing at random (MAR), missing completely at random (MCAR) and non-ignorable data as described in Little and Rubin [9].

## 3 Techniques to Compensate for Missing Data

There are many methods of data fusion for extracting features among variables of highly dimensional data, among these are dimension reduction techniques such as Principal Component Analysis (PCA), Fisher Linear Discriminant (FLD), Multi-dimensional Scaling (MDS), Independent Component Analysis (ICA), Factor Analysis (FA), and Auto associative neural network encoder (autoencoder).

All the above methods except autoencoders are dimension-reduction techniques, in the sense that they can be used to replace a large set of observed variables with a smaller set of new variables, whilst preserving all of the original information. PCA takes advantage of redundancy of information and simplifies data by replacing a group of variables with a fewer new variables, called principal components. Each principal component is a linear combination of the original variables. All the principal components are orthogonal to each other so there is no redundant information. Everitt and Dunn [10] found that FA works well for finding correlations among data, in contrast to PCA which only summarises data using fewer dimensions. MDS is used to detect underlying dimensions that allows one to explain observed similarities or dissimilarities between the investigated objects. MDS is comparable with FA, the only difference being that FA tends to extract more factors than MDS, forcing the user to manually interpret the results; as a result, MDS often yields more useful solutions. Welling and Webber [11] established that ICA searches for directions in data-space, which are independent across all statistical orders. ICA is related to principal component analysis and factor analysis, yet ICA is a much more powerful technique, because ICA is capable of finding the underlying factors or sources when the classic methods such as PCA and FA fail completely. Although PCA finds the minimum number of components that best represents the data, this best representation is in the least square sense and it does not guarantee any usefulness for discrimination. One needs to reduce the dimensionality, under some constraint of maximizing the class discrimination. Maximizing the discrimination can be achieved by increasing the inter-cluster distances and reducing the intra-cluster distances.

These distances are obtained using between and within class scatter matrices through the FLD method. FLD is referred to as multiple discriminant analysis (MDA) when the number of classes of data exceeds two.

Dimension reduction of original complete data can be used to generate a smaller data set, which can then be enlarged into an approximation of the original data set using a prediction neural network (NN). If there is missing data in the large data set, one can ignore the missing variable and use a dimension reduction technique with the available variables and then run the output of the PCA or FLD, etc., through a neural network to regenerate the original complete data set with the approximated variable. The above approximation process has many steps where errors can be introduced. For this reason this paper will evaluate the autoencoders only.

# 4 Autoencoder

Autoencoders use the principle of contractive mapping to locate a point of convergence given a set of known sensors' data and some unknown sensor data. Contractive mapping occurs when the output distance between two points is less than the input distance between the same points. Mathematically it is a mapping O: X→X on a complete metric space (*X, d*) in which, for any *x* and *y* in that space [12]:

$$d(Ox, Oy) \le k \cdot d(x, y) \qquad (1)$$

where $0 \le k \le 1$

Autoencoders are an application of the Banach Fixed-Point Theorem, which states that, if *f* is a contractive mapping, then there exists a *unique* fixed point $x_0$ for which $f(x_0) = x_0$. Moreover, there exists a sequence $\{x_n\}$, for which any element $x_{n+1} = f(x_n)$, converges, and that convergent point is $x_o$. Fig. 1 shows the structure of an autoencoder. Thompson, *et al*, [12] successfully applied an autoncoders to restore missing sensors by minimising the error between the missing sensor inputs and outputs and also minimising the error between the entire input pattern and output pattern using both missing and known sensors to achieve a final answer.

The input to the network is two vectors whose total dimension is *n*. By definition the input dimension and the output dimension are the same in an autoencoder, while the hidden layer has a lower dimensionality than the input/output. As a rule the ratio of input to hidden layer neurons is 2:1, as in [13]. The input into the neural network is given by *x*. The first vector, $x_k$, is the set of known sensor values for a given input pattern. The second vector is $x_m$, the set of missing sensor values. The autoencoder that is used in this paper is a feed-forward multilayered perceptron (MLP). The parameter $w_{ijk}$ is the matrix of weights whose $(i, j)^{th}$ element is the weight connecting the $i^{th}$ known sensor value to the $j^{th}$ neuron in the hidden layer; $w_{ijm}$ is the matrix for the missing sensors, $b_1$ is the vector of bias weights for the first layer; $w_{kjk}$ and $w_{kjm}$ are the weights on the output. Within the input layer the input is transformed from *x* into *a* using the weights ($w_{ji*}$) and baises ($b_j$) [14].

$$a_j = \sum_{i=1}^{d} w_{ji*} \cdot x_i + b_j \qquad (2)$$

Within the hidden layer the input is further transformed using an activation function, such as:

$$z_j = \tanh(a_j) = \left[\frac{e^{c \cdot a_j} - e^{-c \cdot a_j}}{e^{c \cdot a_j} + e^{-c \cdot a_j}}\right] \qquad (3)$$

The variable c is a constant term. In the output layer the input is further transformed into a variable which can be optimised using weights.

$$a_k = \sum_{j=1}^{m} w_{kj*} \cdot z_j + b_k \qquad (4)$$

At the output stage the variable is passed through another activation function.

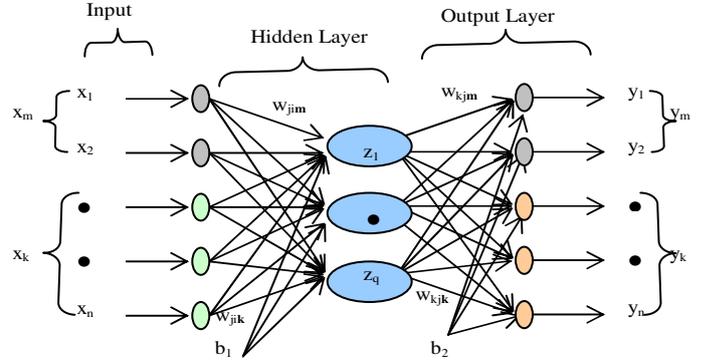

**Fig. 1** Autoencoder architecture

In the case of this work, the activation function that produced the most accurate results was the sigmoid function, shown in (5).

$$y_k = \frac{1}{1 + \exp(-a_k)} \qquad (5)$$

The error at the output of the neural network is given as:

$$e_i(k) = y_j(k) - t_j(k) \qquad (6)$$

Where $t_j$ is the desired output, $y_j$ is the neuron output, and k is the $k^{th}$ output. Within the neural network several methods of optimisation were used to minimise errors in the weights. The optimisation methods tested within the neural networks included gradient methods such as conjugate gradient (CG), scaled conjugate gradient (SCG), quasi-Newton (QN), batch gradient descent (GD). Each stage of the optimisation introduces some error as highlighted in Dhlamini and Marwala [15]. The sum of the squared output error is used to prevent the error from being a negative number, i.e. when target is greater than NN output. The sum squared error is given by:

$$\varepsilon(k) = \frac{1}{2} \sum_{j=1}^{n} e_j^2(k) \qquad (7)$$

Where n is the number of neurons in the output layer. The average squared error is calculated by summing the squared error ($\varepsilon_i$) of all the outputs and dividing by the size of the output set (N), giving an average error ($\varepsilon_{av}$).

$$\varepsilon_{av} = \frac{1}{N} \sum_{j=1}^{n} \varepsilon(k) \qquad (8)$$

Two popular methods of evaluating the error function in the neural network are the maximum likelihood approach and Bayesian training [16]. If there is missing data, then the error function of the autoencoder becomes ($e_m$).

$$e_m = \left( \begin{Bmatrix} x_k \\ x_m \end{Bmatrix} - f\left( \begin{Bmatrix} x_k \\ x_m \end{Bmatrix}, \begin{Bmatrix} w_k \\ w_m \end{Bmatrix}, b \right) \right)^2 \qquad (9)$$

Where the subscript *m* stands for missing, and *k* stands for known. Two evolutionary algorithms were used to optimise the error function and these are particle swarm optimisation and genetic algorithms (GA). Because GA seeks to maximise the error function a negative was inserted in the error equation to obtain a minimum value. So the GA error function is given in (10).

$$e_m = -\left( \begin{Bmatrix} x_k \\ x_m \end{Bmatrix} - f\left( \begin{Bmatrix} x_k \\ x_m \end{Bmatrix}, \begin{Bmatrix} w_k \\ w_m \end{Bmatrix}, b \right) \right)^2 \qquad (10)$$

The process diagram of the evolutionary neural network is as shown in Fig. 2. Approximating missing data using iterative optimisation techniques such as PSO and GA is an expectation maximisation (EM) type of approach. Because first step, called the expectation (E) step computes the expected value of the missing variable. And the second step, called the maximization (M) step, substitutes the expected values for the missing data obtained from the E step and then maximizes the likelihood function as if no data were missing to obtain new parameter estimates. The cycle of expectation and maximisation is repeated until the error is within the tolerance or until the number of cycles has been exceeded.

In this work the autoencoder had 10 inputs and outputs with 7 hidden neurons. The number of neurons in the hidden layer of the classifying MLP was optimised to 31, with 10 inputs and 1 output.

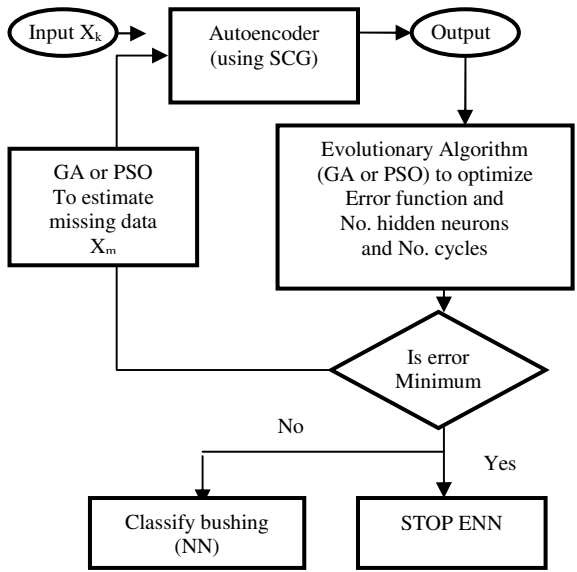

**Fig. 2. Missing data classification flow chart**

## 5 Genetic Algorithms

Genetic algorithms (GA) search the solution space of a function by simulating the survival of the fittest strategy similar to evolution. The fittest individuals of a population reproduce and survive to the next generation. In attempting to simulate evolution GA uses components and stages which include chromosomes, selection functions, genetic functions, reproduction functions, a random initial population, terminating criteria, and an evaluation function [17]. The five stages in a GA optimisation cycle are to create a random initial state, evaluate fitness, select the fittest population, undergo crossover, undergo mutation and repeat until successful. Selection methods include roulette wheel and its variations, scaling techniques, tournament, elitist, and ranking methods. For this work 25 generations and a population size of 20 was used. The roulette wheel was the selection method, arithmetic crossover was used, non-uniform mutation was used. Table 1 compares the results of the simulations done using PSO and GA.

## 6 Particle Swarm Optimisation

Particle swarm optimisation (PSO) takes its origins form the social behaviour of bird flocking. PSO is a random search technique used global optimisation method. There are many similarities between GA and PSO, yet unlike GA, PSO has no evolution operators such as crossover and mutation. In PSO, the potential solutions, called particles, fly through the problem space by following the current optimum particle. Each particle is defined by two variables, a velocity and position, as shown in (11) and (12) respectively.

$$v[\,] = v[\,] + c_1 \cdot rand(\,) \cdot (pbest[\,] - present[\,]) + c_2 \cdot rand(\,) \cdot (gbest[\,] - present[\,]) \qquad (11)$$

$$present[\,] = present[\,] + v[\,] \qquad (12)$$

Where *v[ ]* is the particle velocity, *present[ ]* is the current particle or current solution, *pbest[ ]* is the best solution or fitness achieved so far, *gbest[ ]* is the best value in the global set of particles, *rand ()* is a random number between (0,1) as well as *c1* and *c2* which are learning factors as defined by Shi and Eberhart [18].

Each particle keeps track of its coordinates in the problem space which are associated with the best solution or fitness it has achieved so far. The fitness value is also stored as *pbest*. The best value, obtained so far by any particle in the neighbourhood of the particle is also tracked. This location is called *lbest*. When a particle takes all the population as its topological neighbours, the best value is a global best and is called *gbest*. At each time step, the velocity changes, thus accelerating each particle toward its p*best* and *lbest* locations. Acceleration is weighted by a random term, with separate random numbers being generated for acceleration toward *pbest* and *lbest* locations. Clearly the randomisation of the position of the particles is at the start of the PSO optimisation, i.e. only the first set of particle values are truly random. Subsequent particles all move towards the best particle, *pbest* located at *lbest*. If the initial *pbest* is at a local minimum *lbest* then the entire swarm will search and converge within that local minimum. This is in contrast with GA where the entire search process is random. The starting positions of the chromosomes is random, the mutation of each chromosome is random and further randomised by crossover of mutated chromosomes. When the fittest chromosome is selected, mutated, and reproduced then the search surface is less likely to be a local minimum, because the offspring can end up in a completely unexplored surface, or search the same surface more than once. By remembering the best result over the total number of generations, the GA is able to obtain a global optimum. GA performs a more exhaustive search for an optimum than PSO.

The PSO simulation used a swarm size of 20, with 50 iterations to produce an accuracy of 95% with one missing data point. Table 1 shows more results.

What is common in the implementation of evolutionary techniques like PSO and GA are the following procedures: (1) random generation of an initial population; (2) calculating a fitness value for each subject which is directly dependant on the distance to the optimum; and (3) reproduction of the population based on fitness values.

## 7 Results

Using evolutionary algorithms together with neural networks the work was able to minimise the error function and number of hidden neurons as well as the number of cycles in the iteration. The missing data approximation simulations were done with 500 bushings each with 10 variables. The criterion for a correct approximation of missing value is that it should lie within the standard deviation for each variable, e.g. $CH_2$, and be positive. Standard deviations calculated gave values ranging from 0.96 to 7226.

If the standard deviation is small, the approximated value is closer to the actual missing value. Large standard deviations result in approximated values which are further from the target value even if the approximation is within the standard deviation. Based on the approximated missing variables, 60966 bushings were then evaluated according to IEEEc57.104 and IEC599 criteria, and classified as acceptable or unusable. Table 1 shows the average values of three simulations of the accuracy of approximated values that were calculated using PSO and GA as well as the influence of these estimations on the accuracy of the classification process.

**Table 1. Accuracy of Predictions for Missing Variables and Classification using PSO and GA Key: Est=Estimation; Class=Classification**

| Missing Data | 1 | 2 | 3 | 4 | |
|---|---|---|---|---|---|
| Est. Accuracy | 95% | 84% | 76% | 54% | |
| Class. Accuracy | 96% | 89% | 87% | 79% | |
| Time (s) | 4608 | 4799 | 5006 | 499 | GA |
| Est. Accuracy | 95% | 66% | 68% | 51% | |
| Class. Accuracy | 96% | 64% | 60% | 48% | |
| Time (s) | 1050 | 1057 | 1071 | 1061 | PSO |

PSO was found to be 4 times faster than GA to achieve the same level of accuracy for one missing data point. But it can be argued in that it depends on number of iteration and swarm size. By setting both at very high values, e.g. swarm size of 100 and iterations of 500, the average accuracy of the results can be improved by 1%. To approximate a missing value the *known* data and the guessed unknown/missing value was put into a trained autoencoder. The average error between the input and output values of the *known* values was calculated ignoring the approximated missing variable. If the error in the known variables was greater than $1 \times 10^{-3}$, then the approximated missing value was recalculated by the GA or PSO. The process continued until the number of iterations of the swarm or generations for the GA was exceeded or until the error became less than $1 \times 10^{-3}$. The results obtained shows that an autoencoder can reliably trace correlation between the missing data and known data. The results further show that where 50% of data was not available, the network could only approximate 30% of the missing variables to within the standard deviation.

If PSO and GA performance is compared based on time alone, then PSO is better than GA If accuracy is the criteria for evaluation, and time is not limited then both methods perform the same, because the number of iteration or generations as well as the swarm size or population size can be increased to achieve the desired accuracy. PSO has few parameters to adjust during the optimisation. Furthermore, this figure clearly indicates that as the number of missing values increase the classification accuracy decreases. Furthermore, it indicates that for this particular application it is possible to estimate one missing value and still achieve over 96% classification accuracy. Without any missing value the classification accuracy was 97%. However, the results also indicate that particle swarm optimization as applied in this paper loses accuracy as the number of missing values increase.

## 8 Conclusions

The work finds that an autoencoder can trace correlation between the missing data and known data if 10% of the data is missing both PSO and GA could produce an average accuracy of 95%. Approximation where 100% of missing data was within the standard deviation was also achieved when 10% of the data was missing. If the percentage of missing data is less than 30% of total number variables, the autoencoder approximated the missing data with an average accuracy of 68% for PSO and 76% for GA. PSO was found to be 4 times faster than GA to achieve the same level of accuracy for one missing data point. But it can be argued in that it depends on number of iteration and swarm size. By setting both at very high values, e.g. swarm size of 100 and iterations of 500, the accuracy of the results can improve by 1% but the time to simulate will increased by 400%. GA is found to perform a more exhaustive search for an optimum than the PSO. When the missing data were estimated, it was found that the GA performs better on classifying the condition of the bushings than the PSO.

*References:*
[1] A.J. Mcgrail, E. Gulski, E.R.S. Groot, D. Allan, D. Birtwhistle, T.R. Blackburn, Data Mining Techniques to Assess the Condition of High Voltage Electrical Plant, *Proceedings of Cigre Conference*, Paris, 2002.
[2] D. L. Hall and J. Llinas, editors, *Handbook of Multisensor Data Fusion*, CRC Press, 2001.
[3] J.E. Jackson, *A users guide to Principal Components*, John Wiley and Sons, New York, 1991.
[4] W.G. Madow, I. Olkin, D.B. Rubin, *Incomplete Data in Sample Surveys II: Theory and Bibliographies*, Academic Press Inc., New York, 1983.
[5] M. Abdella, T. Marwala, The use of genetic algorithms and neural networks to approximate missing data in database, *Proceedings of the IEEE 3rd International Conference on Computational Cybernetics*, 2005, Mauritius, pp. 207-212.
[6] M.K. Markey, A Patel, Impact of Missing Data in Training Artificial Neural Networks for Computer-Aided Diagnosis, 2004, website, [www.bme.utexas.edu/research/informatics/pubs/Markey2004_Impact.pdf]
[7] Z. Ghahramani, M.I. Jordan, Mixture models for Learning from incomplete data, Chapter in R. Greiner, T. Petsche, S.J. Hanson, *Computational Learning Theory and Natural Learning Systems, Volume IV*: Making Learning Systems Practical, Cambridge, MA: MIT Press, 1997, pp. 67-85.
[8] V. Tresp, S. Ahmad, R. Neuneier, Training neural networks with deficient data, Chapter in J.D. Cowan, G. Tesauro, J.Alspector, *Advances in Neural Information Processing Systems 6*, San Mateo, CA: Morgan Kaufman, 1994 ,pp. 128-135.
[9] R.J.A. Little, D.B. Rubin, *Statistical analysis with missing data,* New York, Wiley, 1987.
[10] B.S. Everitt, G. Dunn, *Applied Multivariate Data Analysis*, Edward Arnold, London, 1991.
[11] M. Welling, M. Weber, Independent Component Analysis of Incomplete Data, *Proceedings of the 6th joint Symposium on Neural Computation*, UCSD, May 22 1999.
[12] B.B. Thompson, R.J. Marks, M.A. El-Sharkawi, On the contractive nature of autoencoders: Application to sensor restoration, *Proceedings of the IEEE Joint Conference on Neural Networks*, July 2003 Portland.
[13] B.B. Thompson, R.J. Marks, M.A. El-Sharkawi, M.Y. Huang, C. Bunje, Implicit learning in autoencoder novelty assessment, *Proceedings of International Joint Conference on Neural Networks, and IEEE World Congress on Computational Intelligence*, 2002, Honolulu, pp. 2878-2883.
[14] C.N. Bishop, *Neural Networks for Pattern Recognition*. Oxford University, 1995.
[15] S.M. Dhlamini, T. Marwala, Modeling inaccuracies from simulators for HV polymer Bushing, *Proceedings of International Symposium on High Voltage*, 2005, Beijing.
[16] B.C.Vilakazi, P.R. Mautla, E.M. Moloto, T.

Marwala, Bushing condition monitoring using standalone classifiers and committee classifiers, *Proceedings of Pattern Recognition Association of South Africa (PRASA)*, Cape Town, 2005.

[17] C.R. Houck, J.A. Joines, M.G. Kay, A genetic algorithm for function optimisation: A Matlab implementation, *Technical Report NCSU-IE Technical Report 95-09*, North Carolina State University, 1995.

[18] Y. Shi, R.C. Eberhart, Parameter Selection in Swarm Optimisation, *Evolutionary Programing VII*: Proc. EP98, Springer-Verlag, New York, 1998, pp 591-600.